\documentclass[11pt]{article}
\usepackage[preprint]{acl}
\usepackage{times,latexsym}
\usepackage[T1]{fontenc}
\usepackage[utf8]{inputenc}
\usepackage{microtype}

\usepackage{amsmath,amssymb,amsfonts}
\usepackage{graphicx}
\usepackage{textcomp}
\usepackage[table]{xcolor}
\usepackage{booktabs}
\usepackage{url}
\usepackage{xspace}
\usepackage{array}
\usepackage{tabularx}
\usepackage{multirow}
\usepackage{algorithm}
\usepackage{algpseudocode}
\usepackage{marvosym}
\usepackage{placeins}

\newcommand{\ours}{ChronoSynth\xspace}
\newcommand{\op}[1]{\texttt{#1}}
\newcommand{\best}[1]{\textbf{#1}}
\newcommand{\second}[1]{\underline{#1}}
\newcommand{\myparagraph}[1]{\vspace{0.5ex}\noindent\textbf{#1.}\ }
\newcolumntype{Y}{>{\raggedright\arraybackslash}X}

\definecolor{chronoStatic}{HTML}{F6F3FA}
\definecolor{chronoPrompt}{HTML}{FAF5F9}
\definecolor{chronoSynth}{HTML}{F4EFF9}
\definecolor{chronoSup}{HTML}{F5F2FA}
\definecolor{chronoUnsup}{HTML}{FBF6FA}
\definecolor{chronoRelA}{HTML}{F3EDF8}
\definecolor{chronoRelB}{HTML}{F8F5FB}

\definecolor{chronoHeader}{HTML}{F1EBF7}
\definecolor{chronoBand}{HTML}{F4EEF9}
\definecolor{chronoRule}{HTML}{91859D}

\newcommand{\supv}[1]{#1}
\newcommand{\unsupv}[1]{#1}
\newcommand{\synthv}[1]{#1}

\title{ChronoSynth: Constraint-Aware Question Synthesis over Temporal Knowledge Graphs}
\author{Feng Liang\textsuperscript{1}, Weixin Zeng\textsuperscript{1}, Runhao Zhao\textsuperscript{1},
Mingjun Zhou\textsuperscript{1}, Shaoshi Yang\textsuperscript{2}, Xiang Zhao\textsuperscript{1}\thanks{\Letter\ Corresponding author.}\\
\textsuperscript{1}National Key Laboratory of Big Data and Decision, National University of Defense Technology, China\\
\textsuperscript{2}Beijing University of Posts and Telecommunications, China\\
fungloeng@gmail.com, \{zengweixin13, runhaozhao, zhoumingjun24, xiangzhao\}@nudt.edu.cn\\
shaoshi.yang@bupt.edu.cn}

\begin{document}
\maketitle

\begin{abstract}
Knowledge graph question generation (KGQG) generates natural-language questions from structured graph evidence and a target answer. When the input graph is temporal, a valid question must also preserve the answer field, time condition, reference event, and temporal order that determine the answer set. Existing KGQG methods and prompting baselines can produce fluent questions that alter these conditions, causing temporal semantic drift. This paper studies temporal KGQG as a constraint-preserving graph-to-question problem and proposes ChronoSynth, a memory-augmented and constraint-aware generation framework. ChronoSynth first constructs a Temporal Intent State to make the intended question semantics explicit. It then retrieves compatible typed question patterns from an indexed pattern memory, adapts them to the input graph, and validates the generated question. Validation feedback guides revisions when a supported inconsistency is detected. Experiments on CRONQUESTIONS and MultiTQ demonstrate state-of-the-art generation performance, with ablations supporting the contributions of its components.
\end{abstract}

\par

\section{Introduction}
Knowledge graph question generation (KGQG) aims to generate natural-language questions from structured graph evidence, often conditioned on a local subgraph and a target answer~\cite{chen2019reinforcement}. Recent database and graph-data research has advanced knowledge-graph querying and temporal graph processing~\cite{SIGMOD02,tspg}. KGQG supports the construction of question-answer pairs and evaluation benchmarks for graph-based question answering~\cite{xiong2022autoqgs,guo2022dsm,zhao2024roleagentqg}, and it also helps expand the coverage and diversity of questions over the same graph semantics~\cite{ren2025r2dqg}.

\begin{figure*}[!t]
\centering
\includegraphics[width=\textwidth]{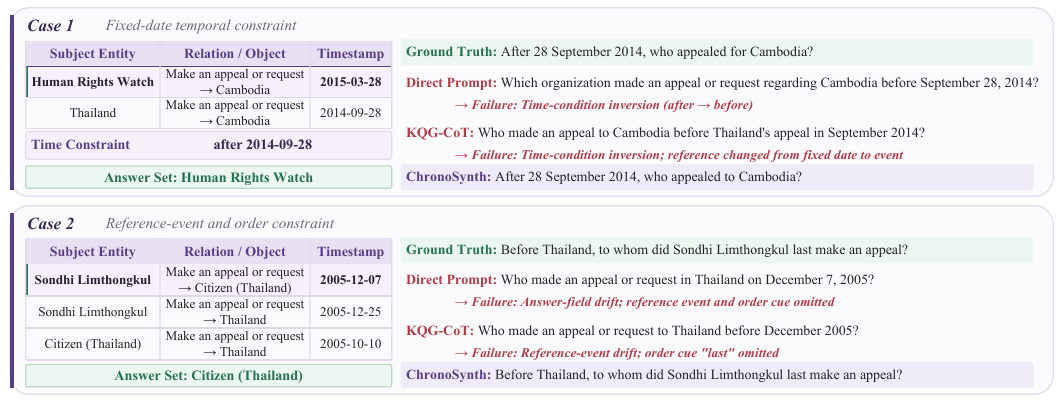}
\caption{Motivating examples of temporal semantic drift. The prompting baselines are fluent but change the time condition, reference event, answer field, or order cue.}
\label{fig:complexity}
\end{figure*}

Temporal knowledge graph question generation (TKGQG) extends KGQG from static triples to time-scoped facts. In a temporal knowledge graph (TKG), a fact may be valid at a timestamp, over an interval, or in relation to another event. This temporal structure changes what a generated question needs to express. Two questions involving the same entities and relation may have different answers because one asks ``before 2014'', another asks ``after 2014'', or one asks for the ``last'' event before a reference event. Changing the answer field, reference point, or temporal order can change the answer set even when the wording remains fluent, as illustrated in Figure~\ref{fig:complexity}. Temporal knowledge graph question answering (TKGQA) benchmarks such as CRONQUESTIONS and MultiTQ make this dependence explicit through questions over timestamps, intervals, and event ordering~\cite{saxena2021cronquestions,chen2023multitq}. TKGQG must preserve the temporal condition that determines the answer. We refer to this requirement as temporal faithfulness.

Despite the importance of TKGQG, generating questions that preserve temporal constraints remains challenging. Closely related KGQG methods generate questions through graph encoding, pattern transfer, or reasoning and refinement with large language models (LLMs)~\cite{chen2019reinforcement,chen2023toward,guo2022dsm,xiong2022autoqgs,liang2023kqgcot,guo2024sgsh,zhao2024roleagentqg,ren2025r2dqg}. However, preserving entities and relations does not ensure that a question retains the temporal conditions governing its answer. As shown in Figure~\ref{fig:complexity}, prompting baselines can reverse a time condition or omit a reference event or order cue despite producing fluent questions. We call this temporal semantic drift: the generated question departs from the intended temporal query and may change its answer set.

TKGQG requires addressing three challenges. First, the same temporal facts can support different questions, requiring the intended answer field and temporal rule to be identified from the input graph, constraints, and target answers. Second, similar relation patterns can express different temporal requirements, so pattern reuse must account for temporal compatibility. Third, generation can alter these requirements or disclose the answer, making output validation necessary even when the question is fluent.

This paper studies TKGQG as a constraint-preserving graph-to-question generation problem. Given a local temporal subgraph and an answer set, the goal is to generate one answerable question that preserves the temporal intent of the input instance. Our central idea is to make this intent explicit before question realization and use it to retrieve compatible typed question patterns. Under this view, temporal rules are semantic constraints rather than optional surface cues.

To address these challenges, we propose \ours, a memory-augmented and constraint-aware framework for TKGQG. It constructs a Temporal Intent State to specify the intended question, builds an indexed pattern memory from training questions, and retrieves patterns by temporal compatibility. Typed adaptation binds the patterns to the input graph, while validation and feedback refinement check the generated question. Experiments on CRONQUESTIONS and MultiTQ show improved generation quality across both datasets and generator backends.

We make the following contributions.
\begin{itemize}
\item We formulate TKGQG as a constraint-preserving graph-to-question problem and introduce a Temporal Intent State that makes the answer field, temporal rule, reference point, and graph context explicit.
\item We propose \ours, which uses temporal intent to retrieve and adapt reusable question patterns, followed by constraint validation and feedback refinement.
\item We evaluate \ours on CRONQUESTIONS and MultiTQ, demonstrating stronger generation scores than the compared baselines and examining component contributions, temporal faithfulness, and synthesis efficiency.
\end{itemize}

\section{Preliminaries}
\label{sec:preliminaries}

\myparagraph{Temporal Knowledge Graph}
A temporal knowledge graph is defined as
\begin{equation}
\mathcal{G}=(\mathcal{E},\mathcal{R},\mathcal{T},\mathcal{F}),\quad
\mathcal{F}\subseteq\mathcal{E}\times\mathcal{R}\times\mathcal{E}\times\mathcal{T},
\end{equation}
where $\mathcal{E}$, $\mathcal{R}$, and $\mathcal{T}$ are the entity, relation, and time domains. Each fact $(s,r,o,\tau)\in\mathcal{F}$ states that relation $r$ holds between subject $s$ and object $o$ at timestamp or interval $\tau$.

\myparagraph{Temporal KG Question Generation}
Given an instance $x=(G_x,A_x)$, where $G_x\subseteq\mathcal{G}$ is a local temporal subgraph and $A_x$ is the target answer set, TKGQG generates a natural-language question $q_x$. The question should satisfy three requirements: \textbf{fluency}, as a well-formed question; \textbf{answerability}, with $A_x$ derivable from $G_x$; and \textbf{temporal consistency}, preserving the input's time condition, reference point, temporal order, and answer field. Explicit constraints are supplied as part of the input graph representation. Temporal rules such as \op{before}, \op{after}, \op{first}, and \op{last} govern answer selection. They are semantic constraints: reversing a direction or dropping an extremum condition can change the answer set. The full rule inventory is given in the appendix.

\section{Methodology}
\label{sec:methodology}

\subsection{Overview}
\begin{figure*}[!t]
\centering
\includegraphics[width=\textwidth]{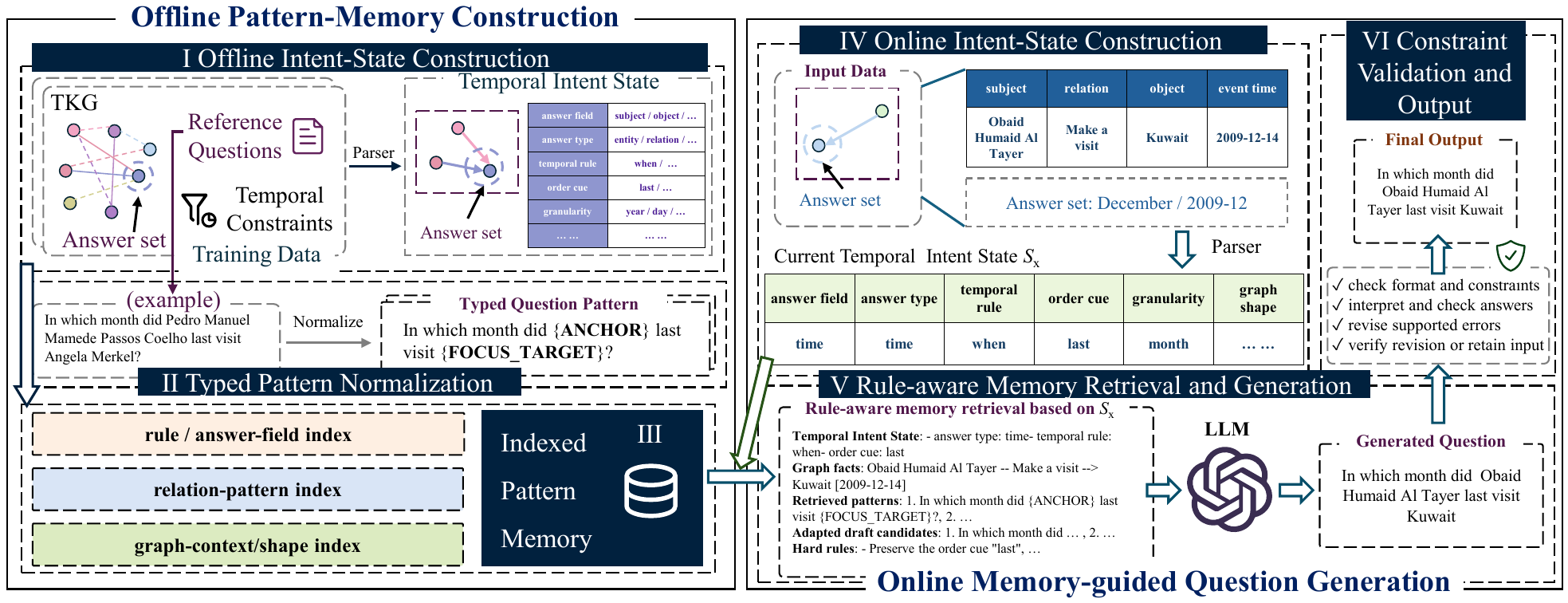}
\caption{Overview of ChronoSynth. ChronoSynth constructs a Temporal Intent State, retrieves compatible typed question patterns from an indexed pattern memory, adapts them to the input graph, and validates and refines the generated question.}
\label{fig:method_overview}
\end{figure*}

\ours addresses temporal semantic drift by making temporal intent explicit before question realization and using it to guide pattern reuse (Figure~\ref{fig:method_overview}). It first constructs a Temporal Intent State from the input graph and answers. An indexed pattern memory supplies training questions and typed patterns with compatible temporal requirements. The generator then realizes a question from these patterns and the input evidence, followed by validation and feedback refinement. Memory construction is performed offline; the same index is reused across generation instances and backends. Algorithm~\ref{alg:chronosynth} summarizes generation using this memory.

\begin{algorithm}[t]
\caption{ChronoSynth question generation}
\label{alg:chronosynth}
\small
\begin{algorithmic}[1]
\Require Graph $G_x$, answers $A_x$, indexed memory $\mathcal M$, generator $p_\theta$, interpreter $R$
\Ensure Generated question $q_x$
\State $S_x \gets \textsc{BuildState}(G_x,A_x)$
\State Retrieve candidates from $\mathcal M$ using $S_x$
\State Rank candidates by Eq.~\ref{eq:retrieval_score}
\State Adapt top patterns to $G_x$ and $S_x$
\State Filter drafts; supplement with default patterns
\State Build context $P_x$ from state, evidence, and patterns
\State $\hat q_x \gets p_\theta(P_x)$
\State $q_0 \gets \textsc{CheckAndRepair}(\hat q_x,S_x,A_x)$
\State $q_x \gets q_0$
\State Interpret $q_0$ with $R$ and execute on $G_x$
\If{a supported inconsistency is found}
    \State Search for a correction preserving $A_x$ and explicit constraints
    \If{a unique minimum-change correction exists}
        \State Express the correction as a revised question $q'$
        \State Reinterpret $q'$ and check the correction, answers, constraints, and format
        \If{all checks pass}
            \State $q_x \gets q'$
        \EndIf
    \EndIf
\EndIf
\State \Return $q_x$
\end{algorithmic}
\end{algorithm}

\subsection{Temporal Intent State Construction}
To select a compatible question pattern, \ours must distinguish the intended answer field, temporal rule, and reference point. We represent these choices and their graph context as
\begin{equation}
S_x=\langle a_x,y_x,\omega_x,g_x,e_x,h_x,c_x,\phi_x\rangle.
\label{eq:intent_state}
\end{equation}
Here $a_x$ is the answer field, $y_x$ the answer type, $\omega_x$ the temporal rule, and $g_x$ the time granularity. The remaining fields specify the focus edge $e_x$, reference entity or event $h_x$, comparison value $c_x$, and graph context $\phi_x$. Together, they distinguish questions that share a relation but ask for different entities, times, or event orders.

The state is inferred from normalized graph records and supplied answers. Explicit constraint records take precedence in determining $\omega_x$. Answer strings and timestamp formats determine $y_x$ and $g_x$. Matching the answers against edge sources, targets, and time fields identifies $a_x$ and the focus edge. The matched temporal pattern then provides the reference point and graph context. Applying this construction to both training instances and new inputs makes their temporal requirements comparable. Field definitions and rule details are provided in the appendix.

\subsection{Indexed Pattern Memory Construction}
\myparagraph{Typed pattern normalization}
For each training instance $z$, \ours replaces instance-specific mentions in its reference question with typed placeholders. The resulting record is
\begin{equation}
m_z=(q_z,\pi_z,k_z,\eta_z),
\end{equation}
where $q_z$ is the reference question, $\pi_z$ the typed pattern, $k_z$ its primary index key, and $\eta_z$ its relation, topology, and surface metadata. Normalization retains the question structure while separating it from particular entities and dates. Anchor, comparison-time, source, and target placeholders preserve distinct semantic roles when the pattern is transferred to a new input.

\myparagraph{Memory organization}
Each record is indexed by fields derived from its Temporal Intent State. The primary key combines the dataset dialect, temporal rule, answer field and type, time granularity, and edge-count bucket. Complementary relation-pattern and graph-shape indexes broaden the candidate pool beyond exact temporal matches; a bounded global pool supplies fallbacks when the combined pool is too small. This organization supports reuse across different entities while retaining the temporal distinctions needed for ranking. The complete index definitions are given in the appendix.

\subsection{Memory-guided Question Generation}
\myparagraph{Rule-aware memory retrieval}
Given $S_x$, \ours collects candidates from the temporal bucket, relation-pattern, and graph-shape indexes, supplementing them from the fallback pool when necessary. It ranks each candidate $m_z$ by a fixed compatibility score:
\begin{equation}
\operatorname{score}(m_z,S_x)
=\sum_{j\in\mathcal J}\lambda_j\kappa_j(m_z,S_x).
\label{eq:retrieval_score}
\end{equation}
Here $\mathcal J$ indexes compatibility features, $\kappa_j$ their values, and $\lambda_j$ their fixed weights. The features measure agreement in temporal rule, answer field and type, granularity, relation, dataset dialect, and graph shape, together with edge-count similarity, reference-token overlap, and reusable placeholders. The temporal rule receives the largest weight because relation-similar instances can require opposite temporal conditions. Relation and graph features provide additional ranking evidence. The weights are fixed before test evaluation. The expanded score and index settings are provided in the appendix.

\myparagraph{Typed pattern adaptation}
For each top-ranked pattern $\pi_z$, \ours constructs a substitution map $\mu_x$ from the current state and graph:
\begin{equation}
\tilde q_{x,z}=\operatorname{Adapt}(\pi_z,\mu_x).
\end{equation}
The map binds anchors, comparison values, and focus entities or times to their respective roles. Keeping these roles distinct prevents relation similarity from making a reference entity interchangeable with an answer entity. \ours discards drafts with unresolved placeholders, literal answer disclosure, or duplicate wording. It retains at most four valid drafts and uses rule-specific default patterns when too few retrieved drafts remain.

\myparagraph{Constraint-aware realization}
The generation context $P_x$ combines the Temporal Intent State, input graph and answers, retrieved memory records, adapted drafts, and generation constraints. A fixed language model produces the initial question:
\begin{equation}
\hat q_x=\arg\max_q p_{\theta}(q\mid P_x).
\end{equation}
Here $\theta$ denotes the fixed generator parameters. The state specifies the intended temporal choices, while the retrieved questions and adapted patterns provide examples of how to express them. The constraints require a single question without unresolved placeholders or answer disclosure and with the intended order cues, temporal rule, and answer field. Generation does not update model parameters or pattern memory.

\subsection{Validation and Refinement}
After realization, \ours cleans the output and checks question format, placeholders, answer leakage, order cues, temporal-rule compatibility, and the answer field. Failed checks trigger the original repair or draft fallback, producing the base question $q_0$.

The final stage checks the meaning of $q_0$ against the input graph, target answers, and explicit temporal constraints. An interpreter maps the question to a temporal query program, which is executed on the graph. When a supported inconsistency is detected, the resulting feedback guides a targeted revision. A second interpretation must confirm that the revised wording expresses the selected correction and preserves the target answers and explicit constraints. If verification fails or no supported correction is available, the base question is retained. This stage refines individual questions without changing the state construction, retrieval procedure, or memory. Detailed indexing, validation checks, and complexity analysis are provided in the appendix.

\FloatBarrier
\begin{table*}[!t]
\arrayrulecolor{chronoRule}
\centering
\footnotesize
\setlength{\tabcolsep}{2.5pt}
\begin{tabularx}{\textwidth}{@{}lYrrrrrr@{}}
\toprule
\multicolumn{8}{c}{\textbf{CRONQUESTIONS}} \\
\midrule
\rowcolor{chronoHeader}
Setting & Method & BLEU-1 & BLEU-2 & BLEU-3 & BLEU-4 & ROUGE-L & CIDEr \\
\midrule
\multirow{4}{*}{\emph{Trained}} & G2S-AE & \supv{0.5011} & \supv{0.3873} & \supv{0.3102} & \supv{0.2577} & \supv{0.4799} & \supv{0.2480} \\
 & G2S-AE-RL & \supv{0.3852} & \supv{0.2628} & \supv{0.1792} & \supv{0.1261} & \supv{0.3850} & \supv{0.0625} \\
 & RoleAgentQG-G2S & \supv{0.3743} & \supv{0.2771} & \supv{0.2123} & \supv{0.1703} & \supv{0.3686} & \supv{0.1037} \\
 & RoleAgentQG-G2S-AE-RL & \supv{0.3842} & \supv{0.2861} & \supv{0.2205} & \supv{0.1778} & \supv{0.3772} & \supv{0.1110} \\
\midrule
\multirow{6}{*}{\emph{Training-free}} & Qwen3.5-9B & \unsupv{0.4237} & \unsupv{0.3122} & \unsupv{0.2417} & \unsupv{0.1952} & \unsupv{0.4029} & \unsupv{1.8729} \\
 & GPT-4o-mini & \unsupv{0.4246} & \unsupv{0.3124} & \unsupv{0.2420} & \unsupv{0.1956} & \unsupv{0.4040} & \unsupv{1.8745} \\
 & KQG-CoT (Qwen3.5-9B) & \unsupv{0.3752} & \unsupv{0.2731} & \unsupv{0.2067} & \unsupv{0.1635} & \unsupv{0.3496} & \unsupv{1.5976} \\
 & KQG-CoT (GPT-4o-mini) & \unsupv{0.4665} & \unsupv{0.3468} & \unsupv{0.2667} & \unsupv{0.2149} & \unsupv{0.4290} & \unsupv{2.0475} \\
\cmidrule(l){2-8}
\rowcolor{chronoSynth}
 & \ours (Qwen3.5-9B) & \synthv{\second{0.5691}} & \synthv{\best{0.4426}} & \synthv{\best{0.3559}} & \synthv{\best{0.2971}} & \synthv{\best{0.5063}} & \synthv{\best{2.7491}} \\
\rowcolor{chronoSynth}
 & \ours (GPT-4o-mini) & \synthv{\best{0.5700}} & \synthv{\second{0.4425}} & \synthv{\second{0.3550}} & \synthv{\second{0.2956}} & \synthv{\second{0.5059}} & \synthv{\second{2.7487}} \\
\midrule
\multicolumn{2}{l}{\textit{Relative improvement}} & +13.7\% & +14.3\% & +14.7\% & +15.3\% & +5.5\% & +34.3\% \\
\bottomrule
\end{tabularx}
\par\medskip
\begin{tabularx}{\textwidth}{@{}lYrrrrrr@{}}
\toprule
\multicolumn{8}{c}{\textbf{MultiTQ}} \\
\midrule
\rowcolor{chronoHeader}
Setting & Method & BLEU-1 & BLEU-2 & BLEU-3 & BLEU-4 & ROUGE-L & CIDEr \\
\midrule
\multirow{4}{*}{\emph{Trained}} & G2S-AE & \supv{0.1509} & \supv{0.0670} & \supv{0.0380} & \supv{0.0234} & \supv{0.1539} & \supv{0.0381} \\
 & G2S-AE-RL & \supv{0.4563} & \supv{0.3510} & \supv{0.2743} & \supv{0.2135} & \supv{0.4799} & \supv{0.4079} \\
 & RoleAgentQG-G2S & \supv{0.1205} & \supv{0.0456} & \supv{0.0210} & \supv{0.0102} & \supv{0.1224} & \supv{0.0099} \\
 & RoleAgentQG-G2S-AE-RL & \supv{0.1537} & \supv{0.0534} & \supv{0.0253} & \supv{0.0130} & \supv{0.1466} & \supv{0.0131} \\
\midrule
\multirow{6}{*}{\emph{Training-free}} & Qwen3.5-9B & \unsupv{0.3977} & \unsupv{0.2683} & \unsupv{0.1881} & \unsupv{0.1345} & \unsupv{0.4088} & \unsupv{1.0367} \\
 & GPT-4o-mini & \unsupv{0.3994} & \unsupv{0.2698} & \unsupv{0.1894} & \unsupv{0.1356} & \unsupv{0.4117} & \unsupv{1.0459} \\
 & KQG-CoT (Qwen3.5-9B) & \unsupv{0.4265} & \unsupv{0.3098} & \unsupv{0.2348} & \unsupv{0.1803} & \unsupv{0.4273} & \unsupv{1.3283} \\
 & KQG-CoT (GPT-4o-mini) & \unsupv{0.4436} & \unsupv{0.3236} & \unsupv{0.2460} & \unsupv{0.1898} & \unsupv{0.4581} & \unsupv{1.5053} \\
\cmidrule(l){2-8}
\rowcolor{chronoSynth}
 & \ours (Qwen3.5-9B) & \synthv{\best{0.5747}} & \synthv{\best{0.4551}} & \synthv{\best{0.3739}} & \synthv{\best{0.3117}} & \synthv{\best{0.5280}} & \synthv{\best{2.6509}} \\
\rowcolor{chronoSynth}
 & \ours (GPT-4o-mini) & \synthv{\second{0.5637}} & \synthv{\second{0.4457}} & \synthv{\second{0.3659}} & \synthv{\second{0.3044}} & \synthv{\second{0.5238}} & \synthv{\second{2.6197}} \\
\midrule
\multicolumn{2}{l}{\textit{Relative improvement}} & +25.9\% & +29.7\% & +36.3\% & +46.0\% & +10.0\% & +76.1\% \\
\bottomrule
\end{tabularx}
\caption{Question-generation results on CRONQUESTIONS and MultiTQ. Bold and underlined values denote the best and second best. Relative improvements compare the best ChronoSynth score with the strongest baseline for each metric.}
\label{tab:full_main_combined}
\end{table*}

\section{Experimental Evaluation}
\label{sec:experiments}

We compare \ours with existing generation methods, examine its components through ablation, and assess question quality and utility through human evaluation and downstream QA. Additional analyses and detailed experimental settings are provided in the appendix.

\subsection{Experimental Setting}

\myparagraph{Datasets}
We evaluate \ours on the graph-converted test splits of CRONQUESTIONS and MultiTQ~\cite{saxena2021cronquestions,chen2023multitq}. CRONQUESTIONS mainly contains timestamp-centered temporal questions. MultiTQ contains more diverse temporal rules and larger local subgraphs. Dataset statistics and additional settings are provided in the appendix.

\myparagraph{Baselines}
We compare \ours with two categories of baselines: parametric trained generation and training-free inference-time generation. The parametric trained group includes graph-to-sequence models and RoleAgentQG variants. The training-free group includes direct prompting, KQG-CoT prompting, and \ours~\cite{chen2023toward,liang2023kqgcot}. In the training-free group, rows named by a backend alone denote direct prompting with that backend. KQG-CoT uses the same backend with reasoning-style prompting, while \ours adds indexed pattern-memory retrieval, typed pattern adaptation, and constraint validation without updating generator parameters. For LLM-based comparisons, we keep the backend family aligned between \ours and its prompting counterparts.

\myparagraph{Metrics}
For the full baseline comparison, we report BLEU-1/2/3/4, ROUGE-L, and CIDEr. BLEU and ROUGE-L measure lexical overlap with reference questions, while CIDEr measures TF--IDF-weighted n-gram similarity. We additionally assess temporal faithfulness and answerability through human evaluation and analyze generation performance across temporal rules, graph sizes, and answer types.

\subsection{Main Results}
Table~\ref{tab:full_main_combined} reports the full-test results on both datasets. It groups baselines by whether they update generation-model parameters. The trained group includes Graph2Seq-style and RoleAgentQG variants. The training-free group includes direct prompting, KQG-CoT, and \ours. Direct prompting and KQG-CoT use prompting only, whereas \ours additionally uses indexed pattern-memory retrieval, typed pattern adaptation, and constraint validation without updating generator parameters. Best values are bolded and second-best values are underlined within each dataset. We organize the results as three observations.

\textbf{Observation 1: \ours consistently improves temporal question generation quality.} On CRONQUESTIONS, \ours reached BLEU-4 0.2971 and CIDEr 2.7491 with Qwen3.5-9B. The GPT-4o-mini configuration obtained BLEU-4 0.2956 and CIDEr 2.7487. By contrast, the best KQG-CoT BLEU-4 was 0.2149, and the best direct-prompting BLEU-4 was 0.1956. Compared with the strongest baseline for each metric, \ours improved BLEU-4 by 15.3\% and CIDEr by 34.3\%. The trained methods were competitive on some lexical metrics, but \ours ranked first and second across all six metrics with its two generator configurations.

\textbf{Observation 2: the improvement is larger on temporally complex MultiTQ questions.} On MultiTQ, \ours reached BLEU-4 0.3117, ROUGE-L 0.5280, and CIDEr 2.6509 with Qwen3.5-9B. With GPT-4o-mini, it obtained 0.3044, 0.5238, and 2.6197, respectively, compared with 0.1898, 0.4581, and 1.5053 for KQG-CoT. The best \ours scores improved BLEU-4 by 46.0\% and CIDEr by 76.1\% over the strongest baseline for each metric. These results indicate that indexed pattern memory is especially useful when the input contains richer temporal rules and larger local subgraphs.

\textbf{Observation 3: the gains extend across generator backends.} \ours improved over direct prompting and KQG-CoT with both Qwen3.5-9B and GPT-4o-mini. These configurations use the same temporal-intent representation, retrieval procedure, and validation policy, with GPT-4o-mini interpreting questions during feedback refinement. The consistent gains support using the framework with either generator.

\subsection{Ablation Study}
\begin{figure*}[!t]
\centering
\includegraphics[width=\textwidth]{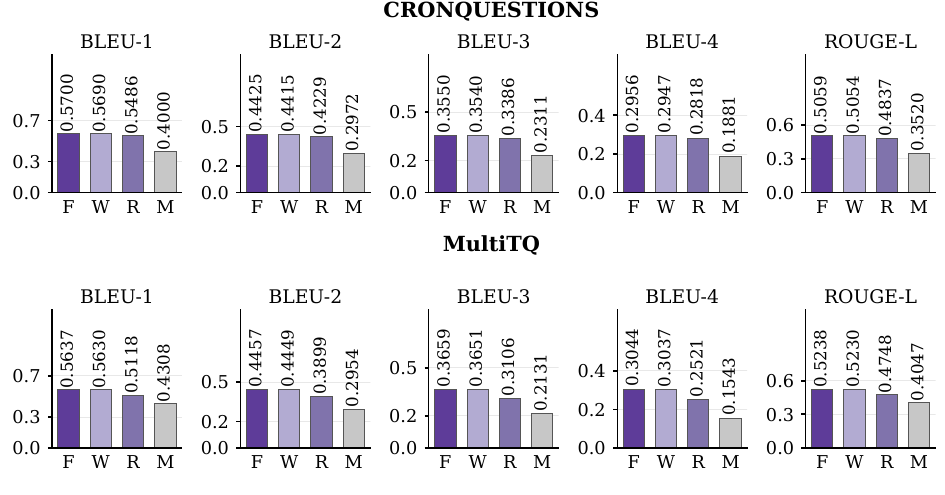}
\caption{Component ablations on CRONQUESTIONS and MultiTQ. F: complete ChronoSynth; W: w/o Refinement; R: Relation-only Retrieval; M: w/o Pattern Memory.}
\label{fig:rq2_ablation}
\end{figure*}

We examine pattern memory, temporal retrieval, and feedback refinement using GPT-4o-mini on the same test splits as the main comparison (Figure~\ref{fig:rq2_ablation}). The variants remove pattern-memory retrieval, replace temporal retrieval with relation-only retrieval, or omit feedback refinement while retaining the original validation checks. The memory and retrieval variants retain feedback refinement. Detailed settings are provided in the appendix.

Pattern memory contributes the largest gain. Removing it lowers BLEU-4 from 0.2956 to 0.1881 on CRONQUESTIONS and from 0.3044 to 0.1543 on MultiTQ. ROUGE-L falls from 0.5059 to 0.3520 and from 0.5238 to 0.4047, respectively. Relation-only retrieval also underperforms the complete method: its BLEU-4 is 0.2818 on CRONQUESTIONS and 0.2521 on MultiTQ. The larger decrease on MultiTQ supports selecting patterns by temporal compatibility rather than relation similarity alone.

Feedback refinement yields smaller, consistent improvements. Without it, BLEU-4 is 0.2947 on CRONQUESTIONS and 0.3037 on MultiTQ. The complete method leads all three variants across BLEU-1 through BLEU-4 and ROUGE-L. The gains indicate that refinement complements the larger contribution of temporal-intent-guided pattern retrieval.

\subsection{Human Evaluation}
We compare questions generated with GPT-4o-mini using direct prompting, KQG-CoT, retrieval-augmented prompting, and \ours. Three raters score fluency, temporal faithfulness, and answerability on a 1--5 scale with method identities hidden. Sampling and annotation details are provided in the appendix.

\begin{table}[tbp]
\arrayrulecolor{chronoRule}
\centering\small
\setlength{\tabcolsep}{3pt}
\begin{tabularx}{\columnwidth}{@{}Yrrr@{}}
\toprule
\rowcolor{chronoHeader}
Method & Fluency & \shortstack{Temporal\\faithfulness} & Answerability\\
\midrule
\multicolumn{4}{l}{\textit{CRONQUESTIONS}}\\
GPT-4o-mini & 2.313 & 2.287 & 2.340\\
KQG-CoT & 2.413 & 2.360 & 2.433\\
Retrieval prompting & 2.380 & 2.360 & 2.373\\
\rowcolor{chronoSynth}
\ours & \textbf{2.987} & \textbf{2.973} & \textbf{2.967}\\
\midrule
\multicolumn{4}{l}{\textit{MultiTQ}}\\
GPT-4o-mini & 2.313 & 2.420 & 2.453\\
KQG-CoT & 2.353 & 2.360 & 2.400\\
Retrieval prompting & 2.353 & 2.393 & 2.347\\
\rowcolor{chronoSynth}
\ours & \textbf{2.873} & \textbf{2.900} & \textbf{2.927}\\
\bottomrule
\end{tabularx}
\caption{Human evaluation on 50 matched inputs per dataset. Scores average three raters on a 1--5 scale; bold marks the highest mean.}
\label{tab:human_main}
\end{table}

Table~\ref{tab:human_main} shows that \ours receives the highest mean rating in all three dimensions on both datasets. Its temporal-faithfulness scores reach 2.973 on CRONQUESTIONS and 2.900 on MultiTQ, compared with baseline maxima of 2.360 and 2.420. Higher answerability ratings further indicate that the generated questions preserve the conditions needed to recover the target answers. These findings complement the reference-based generation metrics.

\subsection{Downstream Question Answering}
We evaluate whether generated questions provide useful supervision for temporal QA. Keeping training graphs and answer sets fixed, we fine-tune Qwen3.5-0.8B using questions from each method and evaluate it on original held-out questions. Synthetic-data conditions use generated questions only; Gold Reference uses the original training questions. Training and evaluation details are provided in the appendix.

\begin{table}[tbp]
\arrayrulecolor{chronoRule}
\centering\small
\setlength{\tabcolsep}{3pt}
\begin{tabularx}{\columnwidth}{@{}Yrr@{}}
\toprule
\rowcolor{chronoHeader}
Training data & \shortstack{CRON-\\QUESTIONS} & MultiTQ\\
\midrule
Gold Reference & $98.33\pm0.42$ & $92.02\pm0.50$ \\
\midrule
GPT-4o-mini & $92.36\pm2.88$ & $73.36\pm2.31$ \\
KQG-CoT & $95.15\pm1.22$ & $76.90\pm1.54$ \\
Vanilla RAG & $88.30\pm0.64$ & $77.71\pm2.07$ \\
\rowcolor{chronoSynth}
\ours & $\mathbf{95.49\pm2.05}$ & $\mathbf{86.37\pm0.10}$ \\
\bottomrule
\end{tabularx}
\caption{Downstream QA EM (\%) with Qwen3.5-0.8B. Values are mean $\pm$ sample standard deviation across three seeds. Bold marks the best synthetic-data mean.}
\label{tab:downstream_main}
\end{table}

\ours achieves the highest mean EM among synthetic-data conditions on both datasets (Table~\ref{tab:downstream_main}). It exceeds the strongest baseline by 0.34 points on CRONQUESTIONS and 8.66 points on MultiTQ. The CRONQUESTIONS difference is small relative to seed variation, while the larger MultiTQ gain supports the utility of preserving richer temporal conditions during synthesis. The remaining gaps to Gold Reference are 2.84 and 5.65 points, respectively.

\section{Related Work}
\subsection{Knowledge Graph Question Generation}
KGQG methods generated questions from structured graph inputs and a target answer. Early neural work encoded a local subgraph and decoded a question with copy mechanisms~\cite{chen2023toward}. Later methods improved coverage with retrieval or diverse-subgraph modeling~\cite{guo2022dsm,xiong2022autoqgs}. Recent LLM-based methods used prompting and reasoning-style generation~\cite{liang2023kqgcot,guo2024sgsh,zhao2024roleagentqg}, while R2DQG studied diverse realizations of the same semantics~\cite{ren2025r2dqg}. Our setting is stricter because the generated question must preserve temporal rules, reference events or times, and answer fields.
Broader question-generation work contributed answer-aware, pretrained, and controlled generation strategies~\cite{L2A,rnn,zero-shot,transformer,s2018,s20182,k2019,t5,IGND,LFKQG,bi2024difficulty}. Related graph-to-text and KBQA settings connected KGQG to SPARQL verbalization, semantic parsing, and benchmark-centered graph question construction~\cite{lecorve2022sparql,template01,saxena2020embedkgqa,jia2021complex}. 

\subsection{Temporal Question Answering over Knowledge Graphs}
Temporal KGQA provides the closest task context. CRONQUESTIONS organizes questions by temporal complexity~\cite{saxena2021cronquestions}, and MultiTQ stresses multi-granularity temporal questions~\cite{chen2023multitq}. Recent TKGQA work studies temporal retrieval, subgraph reasoning, and LLM-guided reasoning~\cite{chen2024ari,qian2024timer4,gao2024gentkgqa,chen2022subgraph,agarwal2024symkgqa,zhang2024qkgr}. These methods answer temporal questions over temporal graphs; our task works in the inverse direction and has to preserve temporal rules during generation.
Related temporal KG research includes surveys, temporal QA variants, and temporal representation learning over dynamic graphs~\cite{liang2024survey,cai2024tkgsurvey,wang2024llmda,tflex2023}. This line of work motivates making temporal intent explicit rather than leaving it implicit in a linearized graph.

\subsection{Retrieval-guided and Constraint-aware Generation}
Retrieval-augmented generation and self-refinement used retrieved evidence or intermediate feedback to improve generation quality~\cite{guu2020realm,lewis2020rag,yao2023react,asai2024selfrag}. Structured-evidence prompting was also common in KG question answering~\cite{baek2023kaping,zhao2024kgcot}. \ours differs in both objective and retrieval unit: retrieval supports question generation rather than answering, the retrieved objects are typed question patterns instead of passages, and validation checks answer leakage, placeholder resolution, order cues, temporal rules, and answer fields.

\section{Conclusion}
We have presented \ours, a constraint-aware question generation framework for temporal knowledge graphs. By making temporal intent explicit through Temporal Intent State construction, indexed pattern-memory retrieval, typed pattern adaptation, and validation with feedback refinement, \ours has reduced temporal semantic drift during question realization. Experiments on CRONQUESTIONS and MultiTQ have shown consistent gains over parametric trained methods and training-free baselines across multiple backends. Ablation and grouped analyses have shown that pattern memory is the main source of improvement. The local efficiency study further demonstrates its quality--cost advantage over KQG-CoT. Future work will extend the temporal-rule inventory and strengthen temporal-consistency checking for broader query styles.

\section*{Limitations}
ChronoSynth relies on a predefined temporal-rule inventory and heuristic state construction, which may require adaptation to new graph schemas or temporal expressions. Pattern reuse depends on the coverage of the training memory, while verification depends on the question interpreter and does not guarantee semantic correctness. Our evaluation focuses on temporal question-generation benchmarks; performance in other domains and languages remains untested.

\section*{Ethical Considerations}
Generated questions may reproduce biases or factual errors in the source graphs and language models. They should therefore be reviewed before use in downstream datasets, particularly in sensitive applications. Reuse of the source datasets and models should follow their licenses and terms of use.

\bibliography{refs}

\clearpage
\appendix
\setcounter{figure}{0}
\renewcommand{\thefigure}{A\arabic{figure}}
\renewcommand{\theHfigure}{appendix.\arabic{figure}}
\setcounter{table}{0}
\renewcommand{\thetable}{A\arabic{table}}
\renewcommand{\theHtable}{appendix.\arabic{table}}

\setcounter{algorithm}{0}
\renewcommand{\thealgorithm}{A\arabic{algorithm}}

\section{Implementation Details}
\subsection{Temporal Rules}
Table~\ref{tab:operator_inventory} summarizes the deterministic state constructor. Explicit constraints take precedence; otherwise, the focus fact and neighboring facts with the same relation provide a temporal interpretation. When several questions are possible, this selects an interpretation rather than recovering a unique reference wording.

\begin{table}[!t]
\arrayrulecolor{chronoRule}
\centering
\scriptsize
\setlength{\tabcolsep}{3pt}
\begin{tabular}{p{0.22\columnwidth}p{0.32\columnwidth}p{0.37\columnwidth}}
\toprule
\rowcolor{chronoHeader}
Temporal rule & Trigger condition & Semantics \\
\midrule
\op{plain} & entity answer, no explicit temporal comparison & ask a non-extremum factual question \\
\op{when} & time answer, no explicit reference & ask for the event timestamp \\
\op{first} & focus edge has the earliest time in its same-relation family & ask for the earliest event or participant \\
\op{last} & focus edge has the latest time in its same-relation family & ask for the latest event or participant \\
\op{before} & later reference exists & ask for an answer before a reference event \\
\op{before\_last} & later reference and latest valid answer before it & ask for the last answer before a reference event \\
\op{after} & earlier reference exists & ask for an answer after a reference event \\
\op{after\_first} & earlier reference and earliest valid answer after it & ask for the first answer after a reference event \\
\op{before\_time} & explicit before-time constraint exists & ask under an explicit before-time constraint \\
\op{after\_time} & explicit after-time constraint exists & ask under an explicit after-time constraint \\
\bottomrule
\end{tabular}
\caption{Temporal-rule inventory.}
\label{tab:operator_inventory}
\end{table}

For inferred rules, the comparison family contains non-constraint facts with the focus relation and anchor entity. Explicit constraints take precedence. Otherwise, entity-answer inputs first check for a later reference fact, then an earlier reference fact; the nearest applicable fact supplies the reference. When neither relative rule applies and several temporal facts remain, the constructor tests whether the focus time is earliest or latest. These decisions use normalized temporal keys. The execution-based validator uses the interval semantics described below.

\subsection{State Fields and Memory Indexes}
Table~\ref{tab:operatorstate_fields} lists the Temporal Intent State fields. Each training record stores its original question, normalized pattern, primary key, and auxiliary metadata. Typed placeholders include \texttt{\{ANCHOR\}}, \texttt{\{CMP\_ENTITY\}}, \texttt{\{CMP\_TIME\}}, \texttt{\{FOCUS\_SOURCE\}}, \texttt{\{FOCUS\_TARGET\}}, and \texttt{\{FOCUS\_TIME\}}.

\begin{table}[!t]
\arrayrulecolor{chronoRule}
\centering
\scriptsize
\setlength{\tabcolsep}{3pt}
\begin{tabular}{p{0.12\columnwidth}p{0.22\columnwidth}p{0.53\columnwidth}}
\toprule
\rowcolor{chronoHeader}
Field & Meaning & Main use \\
\midrule
$a_x$ & answer field & decides whether the question asks for the source, target, or time field \\
$y_x$ & answer type & constrains the wh-word and surface form \\
$\omega_x$ & temporal rule & controls temporal semantics such as \op{before}, \op{after}, and \op{last} \\
$g_x$ & time granularity & distinguishes year-, month-, and day-level realization \\
$e_x$ & focus edge & identifies the fact to verbalize \\
$h_x$ & reference entity/event & provides the reference point for retrieval and drafting \\
$c_x$ & comparison value & specifies the entity, event, or timestamp used for temporal comparison \\
$\phi_x$ & graph context/shape & summarizes topology for memory-retrieval compatibility \\
\bottomrule
\end{tabular}
\caption{Fields in the Temporal Intent State.}
\label{tab:operatorstate_fields}
\end{table}

The primary memory key is
\begin{equation}
k_z=(d_z,\omega_z,a_z,y_z,g_z,b_z),
\end{equation}
where $d_z$ is the dataset dialect, $\omega_z$ is the temporal rule, $a_z$ is the answer field, $y_z$ is the answer type, $g_z$ is the time granularity, and $b_z$ is the edge-count bucket, defined as the edge count capped at six.

\ours stores each record in three complementary indexes: a rule-aware primary bucket index, a relation-pattern index, and a graph-shape index. A bounded global pool supplies fallback candidates. During retrieval, the primary index supplies exact temporal matches, while the relation and graph-shape indexes provide additional candidates for compatibility ranking. The fallback pool is used when the combined candidate pool is too small.

\subsection{Offline Memory Construction}
The memory is built once from the training split. For each record with a reference question, we construct its Temporal Intent State, normalize entity and time mentions into typed placeholders, and store the question, pattern, and retrieval metadata. Records are inserted into temporal, relation, and graph-shape indexes and a bounded fallback pool. The same memory is reused by the online procedure in Algorithm~\ref{alg:chronosynth}; generation and feedback refinement do not modify it.

\subsection{Retrieval Features and Settings}
Candidates are collected from the three indexes. If fewer than 24 candidates are available, the fallback pool supplies additional candidates. Each index bucket stores at most 192 records, and the fallback pool stores at most 512. Candidates are deduplicated by question ID and ranked using the following expansion of Eq.~\ref{eq:retrieval_score}:
\begin{equation}
\begin{aligned}
\operatorname{score}(m_z,S_x)={}&2I_d+4I_\omega+1.5I_f\\
 &+3I_a+2.5I_y+1.5I_g\\
 &+3.5I_r+2I_\phi\\
 &+\delta_{\mathrm{edge}}+\delta_{\mathrm{anchor}}+\delta_{\mathrm{ph}}.
\end{aligned}
\label{eq:retrieval_score_expanded}
\end{equation}
Here $I_d,I_\omega,I_a,I_y,I_g,I_r,I_\phi$ indicate agreement in dataset dialect, temporal rule, answer field, answer type, granularity, relation pattern, and graph shape. $I_f$ is one only when the temporal rules differ but share the same prefix before an underscore; it is zero for exact rule matches. Thus, exact and partial rule agreement are not counted together. The additional terms are
\begin{align}
\delta_{\mathrm{edge}} &= 1/(1+|b_z-b_x|),\\
\delta_{\mathrm{anchor}} &= 0.4\min(|H_z\cap H_x|,3),\\
\delta_{\mathrm{ph}} &= 0.2\min(P_z,4),
\end{align}
where $b$ denotes the edge-count bucket, $H$ contains normalized anchor tokens, and $P_z$ counts typed placeholders. These are fixed implementation weights. Edge-count bucket and string-form question ID break score ties. Each prompt retains four memory records and at most four valid adapted drafts. Rule-specific drafts supplement insufficient valid adaptations.

\subsection{Validation and Feedback Refinement}
Initial checks cover question format, unresolved placeholders, literal answer disclosure, temporal cues, and the requested answer field. Cleaning, repair, or draft fallback produces the base question. The interpreter then receives only this question and its local graph, without reference questions, the target answer list, or the inferred state. Its recovered program is executed and compared with the target answers and explicit input constraints.

\paragraph{Program representation.}
The program schema consists of \op{match}, \op{filter}, \op{extremum}, \op{project}, \op{time\_boundary}, and \op{precision}. These fields specify which facts to match, which temporal conditions to apply, and which answer field to return. Malformed programs, ambiguous references, and unsupported operations receive an unverified status.

\paragraph{Temporal semantics.}
Temporal values denote closed periods at their stated precision. Before requires the fact's end to precede the reference period's start; after requires its start to follow the reference period's end. Explicit-date equality uses interval intersection. An event reference must resolve to a distinct period. First and last select the earliest and latest applicable endpoints and retain ties. Projection cannot return finer precision than the source provides. Open-ended and invalid intervals are unsupported.

\paragraph{Correction policy.}
Correction search can change filter direction, first/last selection, interval boundary, or projected time precision. It changes at most two supported temporal fields while keeping entity and relation bindings fixed. Feedback refinement selects a feasible program only if it uniquely minimizes the number of field changes. Execution-checked surface rules may also restore an explicit input date scope or revise first/last wording.

If no surface rule applies, the original generator proposes a rewrite. A second interpretation must recover a program equivalent to the selected target. Its execution must return the target answer set and satisfy supported explicit constraints, and the question must pass format and answer-disclosure checks. Failure of any check leaves the initial question unchanged. Excluding retries, feedback refinement uses at most two interpretations and one model rewrite per input.

\paragraph{Verification criterion.}
Entity bindings, temporal references, extrema, and interval-boundary cues must be grounded in the question. Equivalence is defined over the program schema: equal answers on one graph do not establish equivalent programs. Acceptance therefore requires both realization of the selected correction and consistency with the execution checks.

\subsection{Complexity and Reuse}
For $N$ training records, index construction takes $O(N\bar c)$ time, where $\bar c$ is the average state-construction and pattern-normalization cost. Storage is linear in the total size of retained records and postings. Ranking $C$ candidates takes $O(C\log C)$ time; adapting $k$ selected patterns takes $O(kL)$, where $L$ bounds pattern length. These costs exclude language-model inference and program execution. The memory is built once and reused across inputs, while verification requires at most two interpretations and one model rewrite per question, excluding retries.

\section{Data and Experimental Settings}
\subsection{Benchmark Preparation}
We use the graph-converted splits of CRONQUESTIONS and MultiTQ. The conversion constructs local graph evidence from the source QA records. Generation receives graph facts, target answers, and explicit constraints encoded in the input. Reference wording and source annotations are removed from the generation input. Reference questions are retained separately for evaluation.

\begin{table}[tbp]
\arrayrulecolor{chronoRule}
\centering\small
\begin{tabularx}{\columnwidth}{@{}Yrrr@{}}
\toprule
\rowcolor{chronoHeader}
Dataset & Train & Dev & Test\\
\midrule
CRONQUESTIONS & 212,259 & 19,014 & 19,122\\
MultiTQ & 257,687 & 39,128 & 36,318\\
\bottomrule
\end{tabularx}
\caption{Numbers of graph-converted records.}
\label{tab:dataset_stats}
\end{table}

\subsection{Generation and Scoring}
GPT-4o-mini and Qwen3.5-9B generate and rewrite questions. Both main-table configurations use GPT-4o-mini for program interpretation; the separate efficiency experiment uses only local Qwen inference. GPT-4o-mini runs use temperature zero, a 900-token response limit, and up to 50 concurrent requests. Local Qwen generation is greedy with thinking disabled. Cached responses are reused only when the prompt and model configuration match.

Baseline scores come from the corresponding benchmark runs. RoleAgentQG-G2S denotes the non-RL G2S-AE variant, whose saved predictions are scored separately on each dataset. BLEU uses corpus-level n-gram precision and brevity penalty. ROUGE-L averages question-level scores with $\beta=1.2$. Both use lowercased whitespace tokens. CIDEr uses case-preserving input and its native scale. All test records remain in evaluation, including unresolved outputs.

\subsection{Resources and Intended Use}
We use the graph-converted CRONQUESTIONS and MultiTQ benchmarks for English temporal question generation. Each input contains local graph facts, target answers, and temporal constraints. The supplements provide data formats, split statistics, implementation, and evaluation records for research on question synthesis and temporal question answering. Table~\ref{tab:resource_sources} identifies the main external resources.

\begin{table}[tbp]
\arrayrulecolor{chronoRule}
\centering\footnotesize
\setlength{\tabcolsep}{3pt}
\begin{tabularx}{\columnwidth}{@{}p{0.35\columnwidth}p{0.30\columnwidth}Y@{}}
\toprule
\rowcolor{chronoHeader}
Resource & Use & Source \\
\midrule
CRONQUESTIONS & temporal benchmark & \href{https://github.com/apoorvumang/CronKGQA}{CronKGQA}~\cite{saxena2021cronquestions} \\
MultiTQ & temporal benchmark & \href{https://github.com/czy1999/MultiTQ}{MultiTQ}~\cite{chen2023multitq} \\
Qwen3.5-9B & local generation & \href{https://huggingface.co/Qwen/Qwen3.5-9B}{Qwen model card} \\
Qwen3.5-0.8B & downstream reader & \href{https://huggingface.co/Qwen/Qwen3.5-0.8B}{Qwen model card} \\
GPT-4o-mini & generation and interpretation & OpenAI \\
GPT-4.1-mini & supplementary judge & OpenAI \\
BLEU, ROUGE-L, CIDEr & reference scoring & software supplement \\
\bottomrule
\end{tabularx}
\caption{External resources and their experimental roles.}
\label{tab:resource_sources}
\end{table}

The Qwen model cards specify Apache 2.0 for the two local models. Source datasets remain subject to their original distribution terms, and the supplements preserve source attribution and third-party notices. Model weights are not redistributed. These notices do not grant additional rights over source data or hosted models.

\paragraph{Software environment.}
Downstream reader experiments use PyTorch 2.5.1 and Transformers 5.12.1. Local generation uses vLLM 0.23.0 with Qwen3.5-9B. The software supplement provides dependency requirements, the model configuration, and the scoring implementation.

\section{Ablation and Grouped Results}
\subsection{Ablation Configurations}
The complete method and three variants use the same complete test splits and GPT-4o-mini for generation, interpretation, and rewriting. Outputs are aligned by test position and question ID so that repeated IDs remain distinct. The w/o Refinement variant retains initial validation and omits feedback refinement. Relation-only Retrieval uses relation-based candidate selection and ranking instead of temporal compatibility. The w/o Pattern Memory variant omits retrieved examples but retains state construction and default drafts. The latter two variants retain feedback refinement.

\subsection{Performance across Input Groups}
We regroup the saved final GPT-4o-mini outputs used in the main table and ablation figure. Groups use the temporal rule inferred from the generation input, its edge count, and the answer type. For each group, BLEU-4 is recomputed at corpus level and ROUGE-L is averaged over its questions. Tables~\ref{tab:groups_cq} and~\ref{tab:groups_mt} report every observed group; no new generation is performed. Positive differences favor the complete method over w/o Pattern Memory.

\begin{table}[tbp]
\arrayrulecolor{chronoRule}
\centering\small
\setlength{\tabcolsep}{3pt}
\begin{tabularx}{\columnwidth}{@{}Yrrr@{}}
\toprule
\rowcolor{chronoHeader}
Group & Count & $\Delta$BLEU-4 & $\Delta$ROUGE-L\\
\midrule
\multicolumn{4}{l}{\textit{Temporal rule}}\\
after\_first & 1,494 & +0.2253 & +0.3688\\
before & 27 & +0.2256 & +0.2837\\
before\_last & 1,907 & +0.1861 & +0.1670\\
first & 1,030 & +0.0928 & +0.1244\\
last & 879 & +0.1095 & +0.1355\\
plain & 7,877 & +0.1156 & +0.1974\\
when & 5,908 & +0.0473 & +0.0445\\
\midrule
\multicolumn{4}{l}{\textit{Edge count}}\\
1 & 12,376 & +0.0739 & +0.1324\\
2 & 6,745 & +0.1632 & +0.1934\\
3 & 1 & +0.5157 & +0.3049\\
\midrule
\multicolumn{4}{l}{\textit{Answer type}}\\
entity & 11,502 & +0.1460 & +0.2149\\
time & 7,620 & +0.0457 & +0.0618\\
\bottomrule\end{tabularx}
\caption{Grouped gains on CRONQUESTIONS. Differences are complete ChronoSynth minus w/o Pattern Memory.}
\label{tab:groups_cq}
\end{table}

\begin{table}[tbp]
\arrayrulecolor{chronoRule}
\centering\small
\setlength{\tabcolsep}{3pt}
\begin{tabularx}{\columnwidth}{@{}Yrrr@{}}
\toprule
\rowcolor{chronoHeader}
Group & Count & $\Delta$BLEU-4 & $\Delta$ROUGE-L\\
\midrule
\multicolumn{4}{l}{\textit{Temporal rule}}\\
after & 675 & +0.1627 & +0.1454\\
after\_first & 2,784 & +0.1168 & +0.1160\\
after\_time & 1,331 & +0.1152 & +0.0948\\
before\_last & 8,620 & +0.0898 & +0.0699\\
before\_time & 1,441 & +0.1162 & +0.0906\\
first & 6,293 & +0.2526 & +0.1955\\
last & 3,697 & +0.2407 & +0.1897\\
plain & 4,890 & +0.2165 & +0.1451\\
when & 6,587 & +0.1100 & +0.0610\\
\midrule
\multicolumn{4}{l}{\textit{Edge count}}\\
1 & 5,201 & +0.2069 & +0.1367\\
2 & 14,467 & +0.1560 & +0.1271\\
3 & 13,470 & +0.1354 & +0.1116\\
4 & 2,011 & +0.1170 & +0.0862\\
5 & 1,169 & +0.0973 & +0.0816\\
\midrule
\multicolumn{4}{l}{\textit{Answer type}}\\
entity & 23,608 & +0.1382 & +0.1115\\
time & 12,710 & +0.1708 & +0.1330\\
\bottomrule\end{tabularx}
\caption{Grouped gains on MultiTQ. Differences are complete ChronoSynth minus w/o Pattern Memory.}
\label{tab:groups_mt}
\end{table}

The complete method improves both metrics in every reported group. On CRONQUESTIONS, the BLEU-4 gain is 0.1460 for entity answers and 0.0457 for time answers. On MultiTQ, the corresponding gains are 0.1382 and 0.1708. Improvements therefore extend across answer types and graph sizes. Counts are reported because small groups, especially the single three-edge CRONQUESTIONS record, do not support general conclusions on their own.

\section{Human Evaluation Protocol}
The main-text human evaluation uses matched inputs and a shared annotation rubric. Details of sampling, annotation, and aggregation follow.

\paragraph{Participants.}
Three volunteers were recruited through a fair process and received reasonable compensation for the annotation work. Participation was voluntary, and all volunteers consented to participate in the evaluation.

\paragraph{Annotation instructions.}
Raters assess each question using only the displayed graph facts, target answers, and expected temporal conditions. They do not consult reference questions, infer the generation method, or use outside information. Fluency, temporal faithfulness, and answerability are scored independently from 1 to 5 using Table~\ref{tab:annotation_rubric}. Answerability is assessed against the complete target answer set, and literal answer disclosure lowers the score. The interface accepts three successive scores per item and saves progress when the rater exits. The software supplement includes the English annotation guide.

\begin{table*}[!t]
\arrayrulecolor{chronoRule}
\centering\footnotesize
\setlength{\tabcolsep}{4pt}
\begin{tabularx}{\textwidth}{@{}cYYY@{}}
\toprule
\rowcolor{chronoHeader}
Score & Fluency & Temporal faithfulness & Answerability \\
\midrule
5 & Natural, clear, and grammatical & All required temporal conditions are preserved & Clear question with the correct complete target answer set \\
4 & Minor language issues & Minor issue that does not materially change the temporal meaning & Minor ambiguity \\
3 & Understandable but noticeably awkward & Temporal intent is recognizable but incomplete or ambiguous & Meaningful ambiguity \\
2 & Major language problems & An important temporal component is wrong or missing & Poorly answerable or likely to yield another answer \\
1 & Incoherent or invalid question & Core temporal meaning is reversed, lost, or replaced & Unanswerable, wrong target, or invalid question \\
\bottomrule
\end{tabularx}
\caption{The annotation rubric used in human evaluation.}
\label{tab:annotation_rubric}
\end{table*}

\paragraph{Privacy and use.}
Source entity names are retained because they are required to interpret the graph facts and generated questions. The anonymous supplements omit annotator names and private configuration information. Generated questions should be reviewed before use in sensitive applications.

\paragraph{Sampling and annotation.}
Using seed 20261005, we sampled 50 matched test inputs per dataset. Sampling preserves source-question-type proportions among inputs with available, uniquely aligned predictions from every method. Each input is paired with GPT-4o-mini questions from direct prompting, adapted KQG-CoT, retrieval-augmented prompting, and \ours.

Three raters independently assign scores from 1 to 5 for fluency, temporal faithfulness, and answerability. Each rater receives a separately randomized order with method labels hidden. Items display graph facts, target answers, and temporal conditions derived from source QA metadata, but no reference question. The displayed conditions are independent of the framework's inferred state. Fluency concerns grammatical clarity, temporal faithfulness concerns preservation of the displayed condition, and answerability concerns support for the complete target answer set. Literal answer disclosure is penalized under answerability.

\paragraph{Retrieval comparator.}
The retrieval comparator uses archived predictions from a generic prompt with four training graph--question demonstrations. Relation-token Jaccard similarity ranks the examples, and question ID breaks ties. It uses no temporal state, typed binding, or feedback refinement. The downstream retrieval baseline is generated separately from its specified pool and tie-breaking rule. Human scores are aligned to the exact question text shown to the raters.

Each rater assesses the same 100 \ours questions and 300 baseline questions. For a given input, all methods share the displayed graph, answers, temporal condition, and rubric, allowing paired comparisons on the same evidence.

\paragraph{Aggregation and uncertainty.}
We first average the three ratings for each question, then average questions within each dataset. The overall score weights the three dimensions equally. Paired method differences use 10,000 bootstrap resamples of source inputs and their averaged ratings, with seed 20261005. These are pointwise 95\% intervals without adjustment for multiple comparisons. \ours's mean advantages range from 0.47 to 0.69 points. All 18 intervals have positive lower bounds, the smallest being 0.353.

\section{Downstream Question Answering Protocol}
Each condition uses 1,000 matched training inputs with fixed graphs and answer sets. Synthetic-data conditions use generated questions only; Gold Reference uses the original questions. All conditions share the reader, training budget, and evaluation protocol.

\paragraph{Synthesis pools.}
We sample the synthesis inputs and a disjoint memory pool of 2,000 training records with seed 20261004. Preparation excludes exact held-out question strings and IDs. Retrieval methods share the memory pool. KQG-CoT receives four structurally selected demonstrations and preserves complete answer sets.

Vanilla RAG selects four demonstrations by relation-token Jaccard similarity, breaking ties by memory position. Each demonstration contains a graph and its training question. A generic prompt adds the current graph and target answers and requests one answerable question without answer disclosure. It uses no temporal state, typed adaptation, validation, or feedback refinement. GPT-4o-mini generates one question per input at temperature 0 with a 900-token limit. Retries address failed calls or mismatched returned model identities, not output quality.

\ours uses GPT-4o-mini for generation, program interpretation, and rewriting, retaining unresolved outputs. Corpus audits check that source IDs, graphs, and answer labels match across methods. Shared training resources supply memory, and shared validation data supports checkpoint selection.

\paragraph{Reader optimization.}
All conditions fine-tune the text model of Qwen/Qwen3.5-0.8B with thinking disabled. The prompt contains only the question and graph; it excludes answer labels, templates, and specification fields. Prompt tokens are masked from the loss, and supervision consists of the complete answer set separated by \texttt{|}. We use float32 parameters and AdamW states with BF16 forward computation. The learning rate is $1\times10^{-5}$, weight decay 0.01, effective batch size 32, microbatch size 8, and gradient clipping 1.0. Learning rate decays linearly after 10\% warmup. The trained reader is distinct from the feedback refinement interpreter.

\paragraph{Selection and decoding.}
Each condition is trained for twelve epochs with seeds 42, 43, and 44. We select the checkpoint with the highest EM on a shared set of 1,000 original validation records. Lower validation cross-entropy and then an earlier epoch resolve ties. Input and output limits are 1,024 and 128 tokens, with truncation disallowed. Test decoding is greedy and uses explicit chat and text termination tokens. Five conditions across two datasets and three seeds give 30 runs with equal example and optimizer-step budgets.

\paragraph{Scoring.}
Evaluation covers all 19,122 original CRONQUESTIONS and 36,318 MultiTQ test records. EM requires equality of complete normalized answer sets. F1 averages per-question F1 computed from answer-set precision and recall. Normalization changes case and whitespace while preserving aliases and date precision. We report means and sample standard deviations over three seeds. Every reader answers from supplied local graphs without global retrieval.

\section{Supplementary Model-Based Assessment}
A separate GPT-4.1-mini judge compares the complete method with w/o Pattern Memory on 1,000 matched inputs per dataset, stratified by source-question type with seed 20261008. Both generators use GPT-4o-mini and retain feedback refinement. Method labels are hidden, A/B order is balanced, and each pair receives three independent calls. The judge sees graph evidence, target answers, temporal conditions, and the two questions, but not reference questions.

We average judgments per input and compute paired differences with 10,000 bootstrap resamples. Temporal-faithfulness and answerability gains are 0.474 [0.381, 0.570] and 0.586 [0.485, 0.690] on CRONQUESTIONS, respectively. On MultiTQ, the differences are $-0.165$ [$-0.245$, $-0.086$] and $-0.159$ [$-0.243$, $-0.077$]. Thus, memory improves reference-based generation metrics across both datasets, while this judge's semantic ratings vary by dataset. This comparison uses different samples and a different evaluator from the human study; its scores are not interchangeable with human ratings.

\section{Local Synthesis Efficiency}
\paragraph{Local-model configuration.}
We run \ours, KQG-CoT, Vanilla RAG, and direct prompting on all 19,122 CRONQUESTIONS and 36,318 MultiTQ test inputs. Each method uses Qwen3.5-9B on one H100 GPU with vLLM 0.23.0, BF16 weights, 40 concurrent workers, temperature 0, and a 900-token output limit per request. Thinking and prefix caching are disabled. For \ours, generation, program interpretation, and model-based rewriting all use the local model. This experiment measures the complete local pipeline; the main generation table retains its GPT-4o-mini interpretation configuration.

KQG-CoT uses 12 structure-clustered training demonstrations. Vanilla RAG retrieves four demonstrations from the complete training pool using relation-token Jaccard similarity; ties follow training-memory order. \ours uses the same training pool for its state-guided pattern memory and retains feedback refinement. All methods receive the same graph evidence and target answers. We retain every output, including five empty KQG-CoT predictions on MultiTQ, when computing reference metrics.

\begin{figure*}[!t]
\centering
\includegraphics[width=0.96\textwidth]{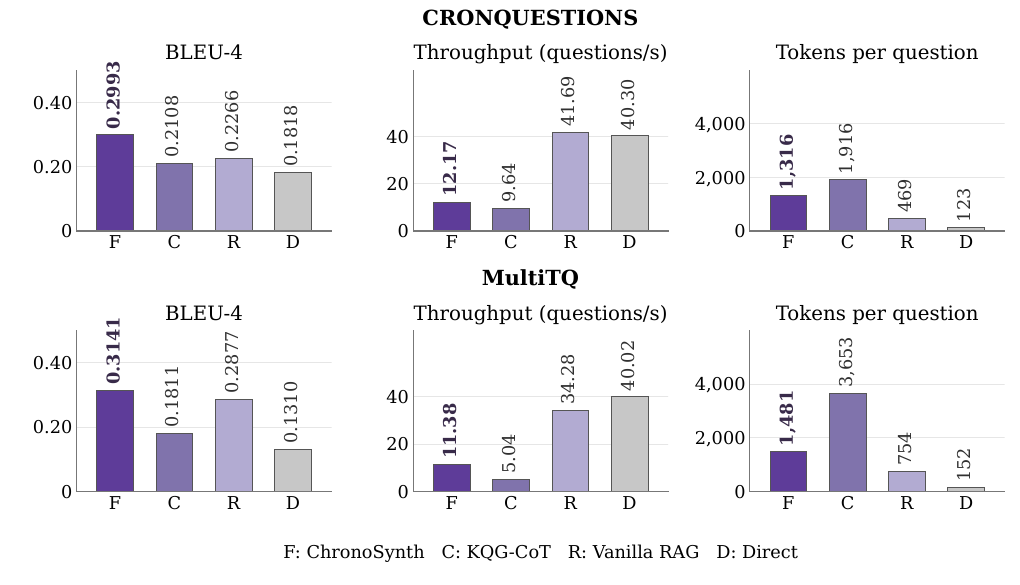}
\caption{Generation quality, throughput, and token use with local Qwen3.5-9B.}
\label{fig:efficiency_overview}
\end{figure*}

\paragraph{Measurement.}
Runtime covers prompt construction and all inference requests, including feedback refinement and retries, after preprocessing and a representative warmup batch. Model startup, data loading, memory construction, and retrieval-index preparation are excluded. Throughput divides the number of test inputs by this runtime. Tokens per question include input and output tokens from all requests. Each method has one complete measured run, so figures report measured values without repeated-run error bars. \ours's preprocessing takes 28.13 seconds on CRONQUESTIONS and 43.57 seconds on MultiTQ.

\paragraph{Recorded inference cost.}
The eight complete local efficiency runs on CRONQUESTIONS and MultiTQ require approximately 4.68 H100 GPU-hours in total. The four ChronoQG runs require approximately 1.64 H100 GPU-hours. Each run uses one GPU. These figures cover the measured inference stages, excluding model startup, preprocessing, and warmup; they do not include reader training, hosted-model calls, or development runs.

\paragraph{Quality and cost.}
Figure~\ref{fig:efficiency_overview} compares generation quality and cost across the four methods. \ours achieves the highest scores on all six reference metrics in this experiment. Its BLEU-4 is 0.2993 on CRONQUESTIONS and 0.3141 on MultiTQ, compared with 0.2108 and 0.1811 for KQG-CoT. \ours completes synthesis in 26.19 and 53.17 minutes, whereas KQG-CoT requires 33.05 and 120.13 minutes. Corresponding throughput is 12.17 and 11.38 questions per second for \ours, and 9.64 and 5.04 for KQG-CoT.

\begin{figure*}[!t]
\centering
\includegraphics[width=0.96\textwidth]{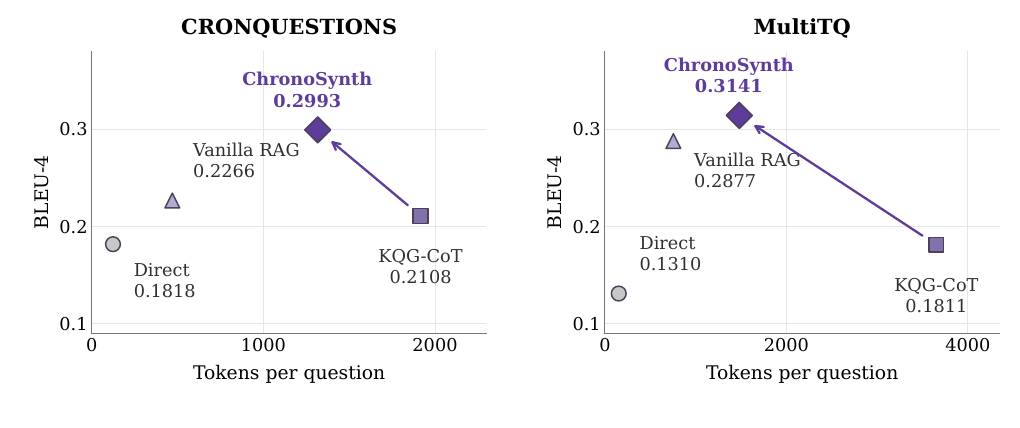}
\caption{BLEU-4 versus total tokens per question. Arrows connect KQG-CoT to \ours.}
\label{fig:efficiency_tradeoff}
\end{figure*}

Figure~\ref{fig:efficiency_tradeoff} shows the quality--token trade-off. \ours uses 1,316 and 1,481 tokens per question, compared with 1,916 and 3,653 for KQG-CoT. This includes an average of 2.12 and 2.18 model requests per question for \ours, versus one for each baseline. Direct prompting and Vanilla RAG require fewer tokens and less runtime, but produce lower BLEU-4. Thus, the efficiency advantage is relative to KQG-CoT, while the quality advantage extends to all three baselines evaluated here.

\paragraph{Output budget.}
Under the common 900-token output limit, KQG-CoT reaches the limit in 126 requests on CRONQUESTIONS and 1,415 on MultiTQ; ChronoSynth does so in zero and seven requests. These outputs remain in evaluation. The comparison therefore measures each method under the stated common decoding budget.

\section{Additional Evaluation on ChronoQG}
\label{app:chronoqg}

\paragraph{Dataset and split.}
ChronoQG-AB85 contains 15,186 temporal graph--question records from CronKG and EventKG, with single- and multiple-constraint questions. Each record supplies graph facts, temporal constraints, a target answer, and a natural-language reference question. As the release has no official training/test split, we use a fixed, approximately 80/20 split with seed 20261011. Records sharing a native source instance or an identical task input are assigned to the same partition, with stratification by source file and temporal-constraint codes. Table~\ref{tab:chronoqg_split} reports the resulting 12,159 training and 3,027 test records.

\begin{table}[tbp]
\arrayrulecolor{chronoRule}
\centering\small
\setlength{\tabcolsep}{3pt}
\begin{tabularx}{\columnwidth}{@{}Yrr@{}}
\toprule
\rowcolor{chronoHeader}
Source subset & Train & Test\\
\midrule
Cron-S & 4,172 & 1,041\\
Cron-M & 1,290 & 309\\
Event-S & 4,955 & 1,236\\
Event-M & 1,742 & 441\\
\midrule
Total & 12,159 & 3,027\\
\bottomrule
\end{tabularx}
\caption{ChronoQG data split. S/M denote single/multiple temporal constraints.}
\label{tab:chronoqg_split}
\end{table}

\begin{table*}[!t]
\arrayrulecolor{chronoRule}
\centering\small
\setlength{\tabcolsep}{3pt}
\begin{tabularx}{\textwidth}{@{}Yrrrrrr@{}}
\toprule
\rowcolor{chronoHeader}
Method & BLEU-1 & BLEU-2 & BLEU-3 & BLEU-4 & ROUGE-L & CIDEr \\
\midrule
Direct Prompting & 0.4710 & 0.3836 & 0.3220 & 0.2761 & 0.4625 & 2.3437 \\
KQG-CoT & 0.3987 & 0.3339 & 0.2892 & 0.2557 & 0.4339 & 2.6055 \\
Vanilla RAG & 0.5625 & 0.4832 & 0.4271 & 0.3834 & 0.5711 & 3.4138 \\
\rowcolor{chronoSynth}
\ours & \textbf{0.6201} & \textbf{0.5495} & \textbf{0.4979} & \textbf{0.4562} & \textbf{0.6472} & \textbf{4.3373} \\
\bottomrule
\end{tabularx}
\caption{Question generation on ChronoQG with local Qwen3.5-9B. Bold marks the highest score.}
\label{tab:chronoqg_quality}
\end{table*}

\paragraph{Adaptation and setting.}
All four methods receive the same graph facts, target answers, and native temporal conditions. The input adapter retains fact identifiers, entity/relation bindings, point/interval types, and time granularity. For \ours, native constraint codes and their conjunctions populate the Temporal Intent State and pattern-memory indexes. Retrieval weights, memory caps, and the generation procedure are unchanged. The validation adapter checks answer-role projections and grounded temporal relation pairs for supported native constraints; unresolved cases retain their initial questions.

Direct prompting and KQG-CoT reuse their generation recipes, with KQG-CoT using 12 structurally selected training demonstrations. Vanilla RAG retrieves four examples from the complete training partition by relation-token Jaccard similarity. \ours builds its pattern memory from the same partition. Every stage uses local Qwen3.5-9B with vLLM 0.23.0, BF16 weights, one H100 per method, and 40 workers. Temperature is zero, the per-request output limit is 900 tokens, and thinking and prefix caching are disabled. Reference questions are used for training examples and offline scoring, and are excluded from test-generation inputs.

\paragraph{Generation results.}
Table~\ref{tab:chronoqg_quality} reports BLEU-1/2/3/4, ROUGE-L, and CIDEr using the same scoring implementation as the main experiments. Scores cover all 3,027 test records, including two empty KQG-CoT outputs. \ours leads the three baselines on all six reference-based metrics. Its BLEU-4 of 0.4562 exceeds Vanilla RAG's 0.3834 by 19.0\%, while ROUGE-L and CIDEr reach 0.6472 and 4.3373. These results extend the generation-quality gains to questions from two source graphs with richer explicit temporal constraints.

\begin{table}[!t]
\arrayrulecolor{chronoRule}
\centering\small
\setlength{\tabcolsep}{3pt}
\begin{tabularx}{\columnwidth}{@{}Yrrr@{}}
\toprule
\rowcolor{chronoHeader}
Method & \shortstack{Time\\(min)} & \shortstack{Tokens\\/question} & \shortstack{Questions\\/second}\\
\midrule
Direct Prompting & 2.53 & 1,333 & 19.91 \\
KQG-CoT & 37.82 & 14,106 & 1.33 \\
Vanilla RAG & 6.49 & 3,682 & 7.77 \\
\rowcolor{chronoSynth}
\ours & 51.65 & 5,012 & 0.98 \\
\bottomrule
\end{tabularx}
\caption{Measured inference cost on ChronoQG.}
\label{tab:chronoqg_cost}
\end{table}

\paragraph{Inference cost.}
Table~\ref{tab:chronoqg_cost} reports one complete measured run per method. Runtime excludes model startup, memory preparation, and warmup; token counts include input and output tokens from every model request. \ours uses 5,012 tokens per question, compared with 14,106 for KQG-CoT, a reduction of 64.5\%. Its additional interpretation and refinement requests yield an average of 2.36 calls per question and a runtime of 51.65 minutes. Direct prompting and Vanilla RAG are faster, while \ours obtains stronger reference-based generation scores.

\end{document}